# Fine-Grained Population Mobility Data-Based Community-Level COVID-19 Prediction Model

Pengyue Jia, Ling Chen∗, Dandan Lyu

College of Computer Science and Technology, Zhejiang University, Hangzhou 310027, China

**Abstract:** Predicting the number of infections in the anti-epidemic process is extremely beneficial to the government in developing anti-epidemic strategies, especially in fine-grained geographic units. Previous works focus on low spatial resolution prediction, e.g., county-level, and preprocess data to the same geographic level, which loses some useful information. In this paper, we propose a fine-grained population mobility data-based model (FGC-COVID) utilizing data of two geographic levels for community-level COVID-19 prediction. We use the population mobility data between Census Block Groups (CBGs), which is a finer-grained geographic level than community, to build the graph and capture the dependencies between CBGs using graph neural networks (GNNs). To mine as finer-grained patterns as possible for prediction, a spatial weighted aggregation module is introduced to aggregate the embeddings of CBGs to community level based on their geographic affiliation and spatial autocorrelation. Extensive experiments on 300 days LA city COVID-19 data indicate our model outperforms existing forecasting models on community-level COVID-19 prediction.

## 1. Introduction

COVID-19 has been spreading globally for more than 2 years since 2019. According to WHO, there have been 196,553k confirmed cases and 4,200k deaths worldwide by the end of July 2021. COVID-19 transmission is still not entirely regulated to this day. As a result, governments and communities have a significant requirement to forecast the number of illnesses with high geographic resolution. On one hand, governments may use prediction data to develop revised anti-epidemic policies (various anti-epidemic measures for locations with varying infection risk levels) and distribution methods (e.g., vaccine distribution) to limit financial losses. Citizens, on the other hand, can use alternative social distances and office habits to decrease infection risk based on prediction information.

Existing models for epidemic prediction fall into three categories. (1) Mechanistic models,

∗ Corresponding author. Tel: +86-13606527774.

*E-mail addresses:* jiapengyue@zju.edu.cn (P. Jia), lingchen@cs.zju.edu.cn (L. Chen), revaludo@zju.edu.cn (D. Lyu).

including compartmental and agent-based models, e.g., Susceptible-Infectious-Recovered (SIR) model. Compartmental models use predefined equations and strict assumptions to predict population-level dynamics, while agent-based models focus on individual contact to simulate the transmission of disease. These models are easy to deploy and can give accurate trend judgments in the early stage of an outbreak. But they are hard to align to the reality and have high computational complexity (Gallagher and Baltimore 2017). (2) Traditional statistical models, e.g., Autoregressive (AR) and Autoregressive Integrated Moving Average (ARIMA) models. These classical time series prediction models have simple analysis process and have great accuracy in short-term prediction. But due to the variability of the influencing factors of infectious diseases, long-term disease propagation patterns are difficult to extract. It is also very hard to pick the right parameters at different stages. (3) Deep learning models, e.g., models based on graph neural networks (GNNs) and recurrent neural networks (RNNs), which combine the temporal and spatial patterns and are natural representations for a wide variety of real-life data (Kapoor et al. 2020). Due to the advantages of deep learning models, GNN-based COVID-19 prediction models are gradually emerging. Some of them firstly aggregate the population mobility data to lower spatial resolution geographic level, e.g., county level, and then construct the graph, which will lose some useful information (Kapoor et al. 2020); while others directly use the coarse-grained population mobility data or simple adjacency to construct the graph (Panagopoulos, Nikolentzos, and Vazirgiannis 2021; Rodriguez et al. 2021). However, most of them pay attention to low spatial resolution prediction, e.g., county-level prediction, which is not sufficient for governments to develop refined anti-epidemic policies, improve material utilization, and decrease the number of infections.

Figure 1 shows an example of the information loss of community-level data aggregation. The number in a node represents the total number of POI visitors in a census block group (CBG) or a community, and an edge represents the population mobility data from an origin CBG/community to a destination CBG/community. Through the example, we can see that there are two types of information loss because of aggregation: (1) CBG connections within a community are ignored. (2) Inter-community connections are simplified.

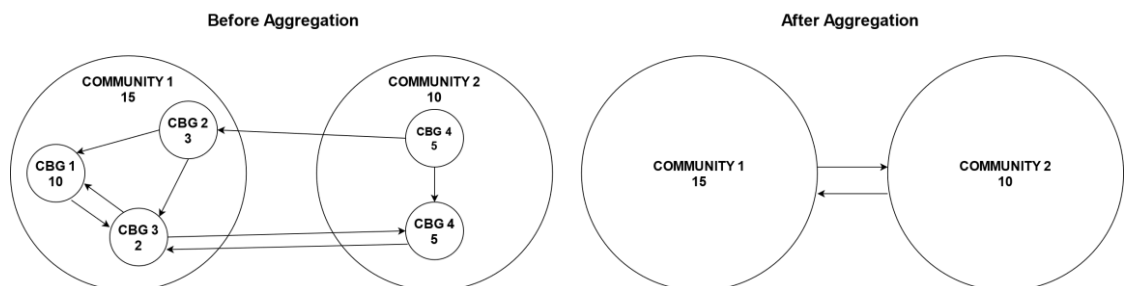

Figure 1: An example shows the information loss of community-level data aggregation

In order to address the aforementioned problems, we propose a fine-grained population mobility data-based community-level COVID-19 prediction model (FGC-COVID), which models the complex spatial patterns by using fine-grained population mobility data that are at the level of individuals, and aggregates CBG embeddings based on geographic affiliation and spatial autocorrelation.

Our contributions are as follows:
1) We build graph on CBG-level. CBGs are the nodes of the graph and each node has three features: POI visits, population, and social vulnerability. In addition, we use fine-grained population mobility data to build edges. Two nodes are connected if there is a population flow between them, and the weight of the edge is determined by the number of population flows. This makes full use of the properties of the connections between low-level geographic units and can decrease information loss and capture the complex spatial patterns.
2) We introduce a spatial weighted aggregation module to aggregate the embeddings of CBGs based on their geographic affiliation to incorporate the spatial distribution patterns of POI visits data. We calculate the local spatial autocorrelation index to obtain the spatial relationship between a CBG and its surrounding CBGs, and take this index as the weight of the geographic affiliation, which can reflect the effect of a CBG on the surrounding environment.
3) We evaluate the model on real COVID-19 data from the city of Los Angeles, comparing it with a broad range of the state-of-the-art models.

The rest of this paper is organized as follows. Section 2 presents an overview of relevant work in COVID-19 prediction. Section 3 describes the details of our proposed model. Section 4 provides our experimental evaluation. Section 5 summarizes our paper and presents our future work.

## 2. Related work

Since the worldwide outbreak of COVID-19, many recent studies have used different models to simulate the spread pattern of COVID-19 and to predict the number of infections or deaths. As mentioned before, there are three main categories of prediction models for epidemic spreading: mechanistic models, traditional statistical models, and deep learning models.

In terms of mechanistic models, SIR, Susceptible-Exposed-Infectious-Recovered (SEIR) models are still very popular choices in epidemic prediction. Recent works have extended the classical model with parameters and structures (Calafiore, Novara, and Possieri 2020; He, Peng, and Sun 2020; Pandey et al. 2020). In addition, the classical SIR, SEIR models need to calculate the constant infection rate at the beginning. To make the models more dynamic, Kiamari et al. (2020) used a time-varying SIR model to predict the trend of COVID-19. Agent-based models (ABM) have also been used by many researchers. These works constructed a complex system by agents following simple rules for predicting the risk of COVID-19 transmission. Cuevas (2020) proposed an agent-based model to evaluate the COVID-19 transmission risks in facilities with defining the mobility requirements and contagion susceptibility of each agent. In addition, Shamil et al. (2021) focused on predicting infections and the impact of intervention, e.g., lockdown and contact tracing, on the spread of COVID-19. Mechanistic models can easily reflect the effect of parameter changes on the results. However, due to the artificially established rules of contagion, these models fail to reflect reality. In addition, the setting of parameters requires huge computing power and is time consuming.

The task of predicting the trend of COVID-19 can be considered as a time series prediction problem, which can be solved by many traditional statistical models. Among them, ARIMA is the most used model because of its high accuracy of prediction (Alzahrani, Aljamaan, Al-Fakih 2020; Ceylan 2020; Roy, Bhunia, and Shit 2021; Singh et al. 2020). Alzahrani, Aljamaan, Al-Fakih (2020) employed the ARIMA model to forecast the daily number of new COVID-19 cases in Saudi Arabia in the following four weeks with comparing to moving average model (MA), AR, and autoregressive integrated moving average model (ARMA). Ceylan (2020) selected the most accurate ARIMA model for predicting the epidemiological trend of COVID-19 in Italy, Spain, and France, by setting different sets of parameters during the early stages of the outbreak. However, due to the instability of epidemic data, e.g., the impact of policy interventions, the prediction results of statistical models can have large deviations.

In terms of deep learning models, on the one hand, temporal models, e.g., Long Short-Term Memory (LSTM) and RNN, are widely used (Arora, Kumar, and Panigrahi 2020; Banerjee and Lian 2022; Jing et al. 2022; Wang et al. 2020; Zeroual et al. 2020). Banerjee and Lian (2022) proposed a novel data driven approach using a LSTM model to form a functional mapping of daily new confirmed cases with mobility data. Jing et al. (2022) proposed a dual-stage attention-based RNN model that combines daily historical time-series data with regional attributes to forecast confirmed COVID-19 cases. On the other hand, due to the interpretability and high prediction accuracy of GNN models in epidemic prediction, a series of GNN-based models have emerged. Structurally, all these models use geographic units as nodes of the graph and construct edges with mobility data or adjacency between nodes. Kapoor et al. (2020) examined a spatio-temporal graph neural network based on infection data and mobility data to predict the number of infections on the US county level. Deng et al. (2020) proposed a graph-based deep learning framework for long-term epidemic prediction from a time-series forecasting perspective. The model captures spatial correlations and temporal dependencies with a dynamic location-aware attention mechanism and a temporal dilated convolution module. Panagopoulos, Nikolentzos, and Vazirgiannis (2021) proposed a model extracting the diffusion patterns and predicting the number of future cases. Due to the limited data, they then used a model-agnostic meta-learning based method to transfer knowledge from one country's model to another. These models have made great progress in prediction accuracy, but the spatial resolution of prediction is mostly at the country and county levels, which is not conducive to fine-grained anti-epidemic policymaking. The difficulty of fine-grained geographic unit prediction is how to capture the dependencies between units and how to aggregate the high spatial resolution embeddings with minimizing information loss. Since units at different geographic levels have strict affiliation, we focus on the bottom units, i.e., CBG-level. In addition, we aggregate the embeddings of nodes upwards based on the affiliation and their spatial distribution patterns, which can reflect the effect of a CBG on the surrounding environment.

## 3. Methodology

In this section, we describe the details of FGC-COVID for forecasting COVID-19 infections. We construct the graph on CBG-level, and our model focuses on community-level COVID-19 prediction.

### 3.1. Definitions

Given the historical COVID-19 data from correlated CBGs and communities they belong to, the task of epidemic prediction is to predict the future infection number of each community.

**CBGs and communities:** We define $CBG = \{cbg_i\}_{i=1}^{N_{cbg}}$ as the set of CBGs, and $COM = \{com_i\}_{i=1}^{N_{com}}$ as the set of communities, where $N_{cbg}$ denotes the number of CBGs and $N_{com}$ denotes the number of communities. One community consists of several CBGs, and one CBG only belongs to one community.

**Aggregation weight:** For $cbg_i$, its aggregation weight is represented as $aw_i$, which controls the aggregation of the embedding of $cbg_i$ to the embeddings of $com_j$, where $cbg_i$ is belonged to $com_j$.

**CBG graph:** Following previous works, we define all correlated CBGs as a weighted directed graph $G = (V, E, A)$, where $V$ is a set of nodes and $|V| = N_{cbg}$, $E$ is a set of edges, and $A \in \mathbb{R}^{N_{cbg} \times N_{cbg}}$ is a weighted adjacency matrix, representing the relationship between nodes. For time step $t$, the graph is represented as $G_t$.

For time step $t$ and $cbg_u$, its node has three features: visits pattern $vst_u^t$, population $pop_u^t$, and social vulnerability score $vul_u^t$, where $pop_u^t$ and $vul_u^t$ are static features, and $vst_u^t$ varies every time step.

For edge $E(u, v)$, its weight is represented as $A_{u,v}$, where $u$ and $v$ are two connected CBGs.

**Infection number:** The infection number of $com_i$ at time step $t$ is represented as $inf_i^t$.

**Historical data:** The historical sequence is denoted as $HD = [hd^1, hd^2, ..., hd^t, ..., hd^T]$, where $T$ is the time window length, $hd^t$ denotes the embeddings at time step $t$, $hd^t = [vst_{1:N_{cbg}}^t, pop_{1:N_{cbg}}^t, vul_{1:N_{cbg}}^t, inf_{1:N_{com}}^t]$.

**Community epidemic forecasting:** Given the historical epidemic data $HD$ and the affiliation of communities and CBGs, community epidemic forecasting task aims to forecast the infection number $[\widehat{inf}^{T+1}, \widehat{inf}^{T+2}, ..., \widehat{inf}^{T+h}] \in \mathbb{R}^{N_{com} \times h}$ between time step $T+1$ and $T+h$, where $h$ denotes the forecast horizon, we set it as 1 in this paper to forecast the next time step infection number. The problem can be formulated as:

$$\widehat{inf}^{T+1} = \mathcal{F}(hd^1, hd^2, ..., hd^t, ..., hd^T; \theta) \quad (1)$$

where $\mathcal{F}$ is the mapping function and $\theta$ denotes all learnable parameters.

### 3.2. Framework

Figure 2 shows the framework of our fine-grained population mobility data-based model. Our model focuses on two geographic levels. Spatial pattern extraction module is on CBG-level utilizing GCN, and temporal pattern extraction module is on community-level. In addition, the conversion of embeddings from CBG-level to community-level is based on the spatial weighted aggregation module.

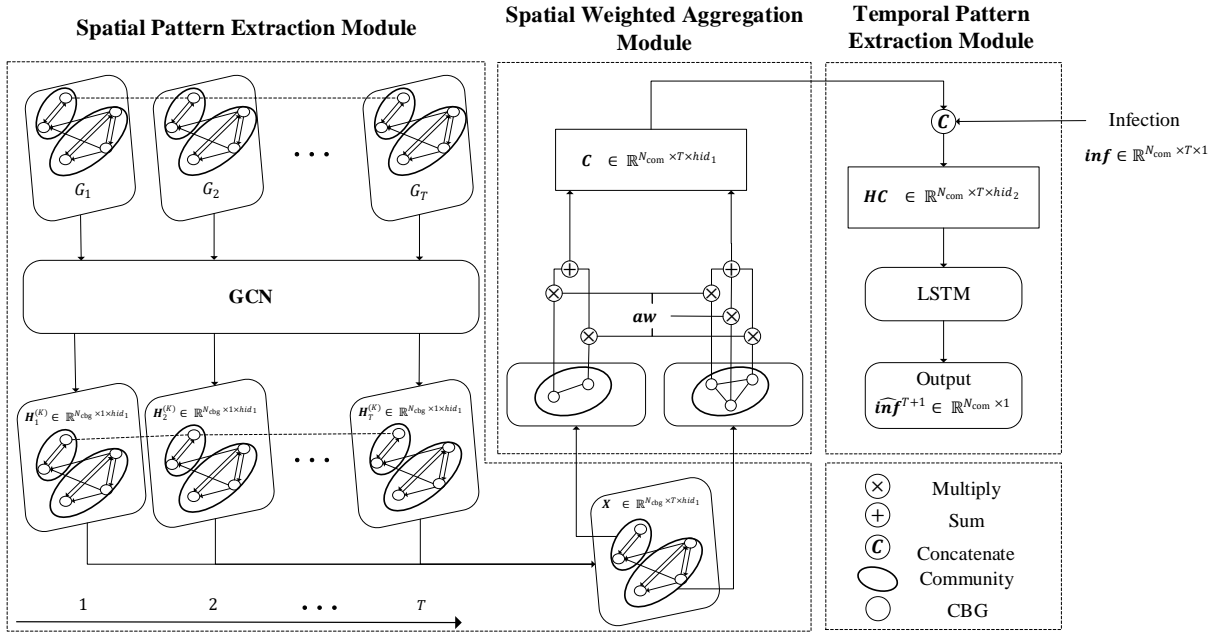

Figure 2: Model architecture.

#### 3.2.1. Spatial Pattern Extraction Module

A spatial pattern extraction module is designed to capture the underlying spatial dependencies at high spatial resolution layer. It takes the CBG-layer data $\{vst^t\}$, $\{pop^t\}$, and $\{vul^t\}$ as input to obtain the spatial representations $X$. The computational process of spatial pattern extraction module is given as follows:

$$b_i^t = mlp(vst_i^t, pop_i^t, vul_i^t) \quad (2)$$

where $b^t$ represents the hidden state of three CBG-level features at time step $t$, trying to exclude the effects of undesirable impacts and to obtain correlations between different characteristics. $mlp$ denotes a function implemented by the multi-layer perceptron (MLP).

The spread of the virus is caused by population mobility. Therefore, on CBG-level, we use GCNConv (Kipf and Welling 2017) to capture spatial patterns between CBGs. For the GCNConv, the adjacency matrix is given as follows:

$$A = \begin{bmatrix} A_{1,1} & \cdots & A_{N_{cbg},1} \\ \vdots & \ddots & \vdots \\ A_{1,N_{cbg}} & \cdots & A_{N_{cbg},N_{cbg}} \end{bmatrix} \quad (3)$$

where $A_{i,j}$ denotes the total number of visitors from $cbg_i$ to $cbg_j$ in the time window. It is different in each time window.

Graph $G_t$ is built based on nodes' feature $b^t$ and the adjacency matrix at time step $t$, which is shown in Figure 3.

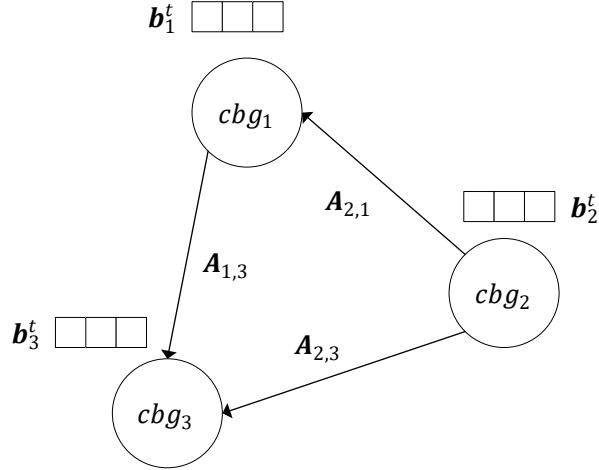

Figure 3: CBG-level Graph $G_t$ at time stamp $t$

We denote the initial CBG embeddings as $h^{(0)}$, and $h_i^{(0)}$ equals $b_i^t$. The GCN process is given as follows:

$$h_i^{(k)} = \gamma^{(k)}\left(h_i^{(k-1)}, AGG_{j\in\mathcal{N}(i)}\phi^{(k)}\left(h_i^{(k-1)}, h_j^{(k-1)}, e_{j,i}\right)\right) \quad (4)$$

where $h_i^{(k)}$ denotes node features of node $i$ in layer $k$, $e_{j,i}$ is the edge features from node $j$ to node $i$. $AGG$ denotes the aggregation function, e.g., sum, mean, or max, and $\gamma, \phi$ are two differentiable functions, e.g., MLP. In addition, we define $H_t^{(K)}$ as the final output of GCN at time step $t$.

The spatial representations are the combination of GCN outputs concatenated in chronological order with the length of the time window:

$$X = \{H_1^{(K)}, H_2^{(K)}, \ldots, H_T^{(K)}\} \quad (5)$$

where $X \in \mathbb{R}^{N_{cbg}\times T\times hid_1}$ and $hid_1$ is a hyperparameter denoting the length of feature after Spatial Pattern Extraction Module.

*3.2.2. Spatial Weighted Aggregation Module*
In order to aggregate CBG-level embeddings to community-level embeddings and reduce information loss as much as possible, we introduce a spatial weighted aggregation module to aggregate the embeddings of CBGs based on their geographic affiliation to incorporate the spatial distribution patterns of POI visits data.

We utilize the Local Moran's I (LMi) (Anselin 1995), which can reflect the effect of a CBG on the surrounding environment, as the weight of the geographic affiliation.

The LMi static of spatial association is given as follows:

$$LMi_i = \frac{x_i - \bar{X}}{S_i^2} \sum_{j=1, j \neq i}^{n} w_{i,j}(x_j - \bar{X}) \tag{6}$$

$$S_i^2 = \frac{\sum_{j=1, j \neq i}^{n}(x_j - \bar{X})^2}{n - 1} \tag{7}$$

where $x_i$ is the sum of visit pattern data in $cbg_i$, $\bar{X}$ is the mean of the $x_i$ in the city, $w_{i,j}$ is the spatial weight between $i$ and $j$ that is calculated by the Inverse Distance, and $n$ is the total number of neighbor CBGs.

We suppose that a CBG with positive LMi has a leading effect on its surrounding CBGs and represents the characteristics of the belonging community. We want to enlarge the spatial pattern of the areas whose visit pattern data form spatial clusters and assign higher weights to them to represent their communities, i.e., assign higher weights to regions that have a positive effect on the cluster formed by their visit pattern data. The positive LMi index represents two types of clusters, i.e., High-High and Low-Low, which means that the region has a positive effect on the clusters they are in. LMi can well represent the spatial associations of visit pattern data with the surrounding CBGs. We use sigmoid, a non-linear function, to normalize LMi index to 0 to 1, which is used as the weight of aggregation:

$$\boldsymbol{aw}_i = sigmoid(LMi_i) \tag{8}$$

The aggregation process is given as follows:

$$\boldsymbol{C}_j = \sum_{i=1}^{N_j} \boldsymbol{aw}_i * \boldsymbol{X}_i \tag{9}$$

where $\boldsymbol{C}_j$ denotes the embedding sequence of $com_j$, $N_j$ is the number of CBGs in $com_j$, $\boldsymbol{aw}_i$ is the aggregation weight of $cbg_i$, and $\boldsymbol{X}_i$ denotes the embedding of $cbg_i$.

### 3.2.3. Temporal Pattern Extraction Module

This module captures the temporal patterns at community level. The input is the embeddings combining the output of spatial weighted aggregation module with the infection number in chronological order. The output of this module is the predicted infection number of each community.

The combination process is given as follows:

$$\boldsymbol{I}^t = mlp(\boldsymbol{inf}^t) \tag{10}$$

$$HI = \{I^1, I^2, \ldots, I^T\} \tag{11}$$

$$C = \{C^1, C^2, \ldots, C^T\} \tag{12}$$

$$HC = cat(HI, C) \tag{13}$$

where $I^t$ denotes the embedding of infection number at time step $t$, $HI$ denotes the embedding sequence of infection number, $HC$ represents the embedding sequence at community level. $cat$ denotes the function that concatenates $HI$ and $C$ in the last dimension. $HC \in \mathbb{R}^{N_{com} \times T \times hid_2}$ and $hid_2$ is a hyperparameter denoting the length of feature after the concatenation in Temporal Pattern Extraction Module.

We utilize a single LSTM to extract temporal patterns, and all communities share the parameters. The prediction process is given as follows:

$$\widehat{inf}_i^{T+1} = mlp(LSTM(HC_i)) \tag{14}$$

where $\widehat{inf}_i^{T+1}$ denotes the predicted infection number of $com_i$ at time step $T+1$. $HC_i$ represents the embedding sequence of $com_i$, whose length is $T$.

We use mean squared error (MSE) to evaluate the errors between true infection numbers and forecasted infection numbers. We define the loss function as follows:

$$L(\theta) = \frac{1}{N_{com}} \sum_{i=1}^{N_{com}} \left| inf_i^{T+1} - \widehat{inf}_i^{T+1} \right|^2 \tag{15}$$

where $N_{com}$ denotes the number of communities, $\widehat{inf}_i^{T+1}$ denotes the predicted infection number of $com_i$ at time step $T+1$, $inf_i^{T+1}$ denotes the true infection number of $com_i$ at time step $T+1$, and $\theta$ denotes all the parameters used in the model.

## 4. Experiment

### 4.1. Datasets

**Geographic relationship between CBG and community**[1]**:** We extract the LA city's CBG list, community list, and their geographic relationship from the Census Block Groups data in LA city geohub.

**Visit pattern**[2]**:** We get the visits data from SafeGraph. The object of this data is POI, and every piece of data has the number and average dwell time of the visitors to a POI. We use their product as the input data. The temporal resolution is a week, but the visitor counts consist of an array with length 7, which means the visitor count's temporal resolution is one day. The left map in Figure 4

---

[1] https://geohub.lacity.org/datasets/lacounty::census-blocks-2020

[2] https://docs.safegraph.com/docs/weekly-patterns

visualizes aggregated Visit pattern on CBG-level.

**Population**[3]: We download population data from the LA city geohub. We use the 2020 population data in this research. The middle map in Figure 4 visualizes Population on CBG-level.

**Vulnerability**[4]: We use the CDC/ATSDR Social Vulnerability Index (SVI) Data 2018. It describes the degree to which a community exhibits certain social conditions, including high poverty, low percentage of vehicle access, or crowded households, which may affect the community's ability to prevent human suffering and financial loss in the event of a disaster. It contains four parts to measure SVI, i.e., Socioeconomic Status, Household Composition & Disability, Minority Status & Language, and Housing Type & Transportation. The right map in Figure 4 visualizes Vulnerability on CBG-level.

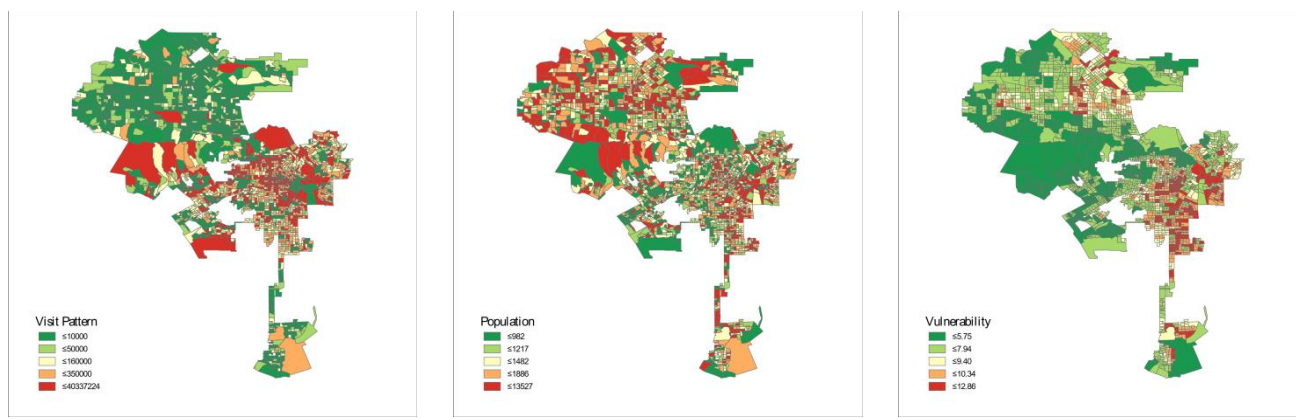

Figure 4: Visit Pattern, Population, and Vulnerability data on CBG-level

**Infection**[5]: We get the daily new case data from GitHub.

**Social distancing metrics**[6]: The data was generated using a panel of GPS pings from anonymous mobile devices. Through this dataset, we capture the mobility between CBGs. However, this dataset is no longer being updated as of 4/19/2021.

The visit pattern, population, vulnerability, infection, and social distancing metrics are collected from June 15, 2020 to April 10, 2021, for a total of 300 days. The numbers of CBGs and communities are 2688 and 139, respectively.

For visits data, we multiply the number of visitors with average dwell time to reflect spread possibility. Because if there are more visitors or more time visitors stay in target locations, there is a greater possibility of virus spreading. While the week's data uploaded on 11/25/2020 are missing, we use the average of the two weeks before and after the week to fill in the missing values.

In addition, to scale the data to the same interval, we use Min-Max Scaler to normalize our data.

---

3 https://geohub.lacity.org/datasets/lacounty::census-blocks-2020

4 https://www.atsdr.cdc.gov/placeandhealth/svi/data_documentation_download.html

5 https://github.com/herf/la-covid

6 https://docs.safegraph.com/docs/social-distancing-metrics

*4.2. Experimental Settings*

Our model is implemented based on PyTorch and PyTorch-geometric Library. Our source code is available on github[7]. All the experiments are conducted on a Linux PC with an Intel Core i9-9900K (8 cores, 3.60G HZ) and NVIDIA RTX 2080Ti.

In our experiment, the input time window size is 21 days, the prediction period is 1 day. A sliding window with a step of 1 day is used to get the samples. We get total 286 samples. We split the datasets into training, validation, and test set in chronological order at the ratio of 50%-20%-30%, which are used to training, validation, and test, respectively. We set the batch size as 32, learning rate as 0.001. The $hid_1$ and $hid_2$ are set to 8 and 24, respectively. In addition, we start early stopping with patience 5 based on validation loss after 150 train epochs. AdamW (Loshchilov and Hutter 2019) is chosen as the optimizer. All experimental results are the average of 5 randomized trials. In our experiment, we use MAE, RMSE, and WMAPE as evaluation metrics.

*4.3. Baselines*

We compare with many classical and up-to-date prediction models to evaluate the benefits of our model.

1) **Autoregressive (AR):** A statistical model that predicts infection number by the linear combination of past data. Each community has an individual model.

2) **LSTM:** A two-layer LSTM model with 32 hidden units and a final output linear layer. The input data are infection data, visits data, vulnerability index, and population data at community level. The linear layer gets the final output from the second LSTM layer and outputs a vector of size 139, which is the community number of LA city.

3) **LSTNET (Lai et al. 2018):** A deep learning model that uses CNN and RNN to extract short-term local dependency patterns among variables and to discover long-term patterns for time series trends. The input data are infection data, visits data, vulnerability index, and population data at community level.

4) **CNNRNN_Res (Wu et al. 2018):** The first deep learning model proposed to predict epidemiology profiles in the time-series perspective. It adopts RNN to capture the long-term correlation and CNN to fuse different datasets. We use the same input data as our model, but with the data resolution of community.

5) **DCRNN (Li et al. 2018):** A graph based deep learning model that captures spatial dependency using bidirectional random walks on the graph, and the temporal dependency using encoder-decoder architecture. Its inputs are the same features as our model at community level.

---

7 Our source code is available on https://github.com/Jia-py/FGC-COVID

6) **STGCN (Yu, Yin, and Zhu 2018):** A spatio-temporal forecasting model that combines graph CNNs and Gated CNNs to capture spatial and temporal patterns.
7) **Google-GNN (Kapoor et al. 2020):** A COVID-19 case prediction model with skip-connections between layers that uses GNN and mobility data.
8) **Cola-GNN (Deng et al. 2020):** A cross-location attention-based graph neural network model that captures temporal patterns in long-term influenza predictions.
9) **MPNN+LSTM (Panagopoulos, Nikolentzos, and Vazirgiannis 2021):** A COVID-19 prediction model that encodes the underlying diffusion patterns that govern the spread into the learning model. Message passing neural networks (MPNNs) are used to update the representations of the vertices of each of the input graphs. In addition, LSTM is used to capture the long-range temporal dependencies in time series.

### *4.4. Results and Analysis*

Table I

The evaluation results of different methods (mean±std).

Boldface indicators the best result and underlined the second-best.

| Model | MAE | RMSE | WMAPE |
|---|---|---|---|
| AR | 4.698±0.325 | 9.672±0.793 | 0.543±0.037 |
| LSTM | 4.422±0.285 | 10.098±0.632 | 0.537±0.035 |
| LSTNET | 5.238±0.616 | 11.633±1.648 | 0.606±0.071 |
| CNNRNN_Res | 4.669±0.255 | 9.870±0.898 | 0.540±0.030 |
| DCRNN | <u>4.083±0.067</u> | <u>8.921±0.296</u> | <u>0.472±0.008</u> |
| STGCN | 5.647±0.698 | 12.599±1.548 | 0.661±0.073 |
| Google-GNN | 5.612±0.503 | 11.000±0.734 | 0.681±0.061 |
| Cola-GNN | 4.780±0.209 | 10.976±1.232 | 0.553±0.025 |
| MPNN+LSTM | 5.092±0.281 | 10.713±0.836 | 0.618±0.034 |
| FGC-COVID | **3.790±0.035** | **8.101±0.117** | **0.439±0.006** |

Table I presents the evaluation results of our proposed model and baselines on LA city COVID-19 dataset. We observe the following phenomena:

1) Our model achieves the best performance for forecasting infection numbers. It outperforms classical epidemic prediction models distinctly and excels the graph neural networks-based prediction models dramatically. It captures mobility patterns from a lower geographic level unit. On one hand, lower geographic levels have higher geographic resolution and are a source of more accurate information. On the other hand, lower geographic levels have more geographic units, allowing the model to have higher fault tolerance. In addition, the model utilizes two modules to capture the spatial and temporal patterns respectively, which integrates both spatial and temporal dimensions. Compared to the second-best model, our model gets a gain of 0.293 on MAE, 0.82 on RMSE, and 0.033 on WMAPE.

2) Among the other models, DCRNN works better. DCRNN is trained by maximizing the likelihood of generating the target future time series using backpropagation through time and is able to capture the spatiotemporal dependencies. Cola-GNN performs poorly in this experiment. The possible reason is that Cola-GNN focuses on the long-term prediction problem, which is different from our work. Another spatio-temporal model, STGCN, does not perform well on this task. It might be because the complexity of STGCN and the limited size of the dataset.

*4.5. Ablation Test*

To analyze the effectiveness of each component in our model, one of these modules is removed or modified at a time. We perform the ablation tests on the same dataset with the same parameters.

- **FGC-COVID w/o cst:** It modifies the edge attributes of the graph to be static. We use the sum of the mobility data between two CBGs in the training set as the edge attribute.
- **FGC-COVID w/o swa:** It removes the spatial weighted aggregation module from our model and directly builds the graph at community level. Each community is a node, and the sum of the mobility data between two communities in the training set is the edge attribute. As the same as FGC-COVID, the embedding after the GCN is spliced with the infection data and fed into the LSTM model.
- **FGC-COVID w/o ewa:** It modifies the weight of each CBG to be equal. We use the sum of the CBG embeddings without weights to represent the community embedding.
- **FGC-COVID w/o twa:** It modifies the weight of each CBG to be trainable.

Table II

The performance of FGC-COVID and its variants (mean $\pm$ std).

| Metrics | FGC-COVID | F w/o cst | F w/o swa | F w/o ewa | F w/o twa |
|---|---|---|---|---|---|
| MAE | 3.790±0.035 | 4.655±0.160 | 4.883±0.375 | 4.033±0.284 | 4.490±0.397 |
| RMSE | 8.101±0.117 | 8.558±0.598 | 9.418±0.845 | 8.357±0.462 | 9.025±0.509 |
| WMAPE | 0.439±0.006 | 0.558±0.008 | 0.604±0.054 | 0.474±0.019 | 0.545±0.048 |

Table II shows the performance of our model and its variants, and we can derive the following analytical results:

(1) FGC-COVID performs better than F w/o swa especially on RMSE. The reason might be that the spatial weighted aggregation module connects different geographic level units and aggregates them to higher-level units, which makes the embeddings of communities keep more useful information for the target task.

(2) FGC-COVID outperforms F w/o cst on all metrics, indicating that dynamic edge attributes for each time window is an effective way to help spatial pattern extraction module capturing spatial connections. The reason might be that more information is introduced in time dimension using dynamic mobility data in each time window.

(3) FGC-COVID outperforms F w/o ewa and F w/o twa on all metrics, illustrating that the spatial weighted aggregation module helps the aggregation process to extract more representative community embeddings. The reason might be that we use the Local Spatial Autocorrelation index, which reflects the effect of a CBG on the surrounding environment, as the weight of the geographic affiliation. In addition, directly parameterizing the weight of the geographic affiliation would introduce more parameters and cause the problem of overfitting.

*Parameter sensitivity*

In this section, we focus on the size of time window and the size of graph features.

Table III

The performance with different sizes of time window (mean±std).

| The sizes of time window | 7 | 14 | 21 | 28 |
|---|---|---|---|---|
| MAE | 4.539±0.237 | 4.190±0.372 | 3.790±0.035 | 5.190±1.523 |
| RMSE | 10.732±0.553 | 8.433±0.167 | 8.101±0.117 | 10.550±1.933 |
| WMAPE | 0.465±0.024 | 0.478±0.029 | 0.439±0.006 | 0.710±0.208 |

Table III shows the performance of our model with different sizes of time window. Under different time window conditions, the best prediction is achieved with a time window of 21 days. The possible reasons for this are that the smaller time window lacks sufficient data for prediction, but bigger time window introduces more parameters and is prone to overfitting due to limited training data.

Table IV

The performance with different dimension numbers of $hid_1$ (mean±std).

| The values of $hid_1$ | 2 | 4 | 8 | 12 |
|---|---|---|---|---|
| MAE | 3981±0.126 | 4.038±0.316 | 3.790±0.035 | 3.931±0.084 |
| RMSE | 8.443±0.167 | 8.356±0.691 | 8.101±0.117 | 8.137±0.183 |
| WMAPE | 0.478±0.029 | 0.467±0.037 | 0.439±0.006 | 0.462±0.018 |

Table IV shows the performance of our model with different dimension numbers of GCN outputs, i.e., $hid_1$. The results indicate that as the value of $hid_1$ increases, the performance of the model is relatively improved and then decreased. The best size of $hid_1$ is 8. The possible reason is that smaller dimension loses information, but too large dimension fails to capture patterns.

Table V

The performance with different dimension numbers of $hid_2$ (mean±std).

| The values of $hid_2$ | 12 | 24 | 36 | 48 |
|---|---|---|---|---|
| MAE | 5.120±0.312 | 3.790±0.035 | 4.272±0.369 | 5.15±0.647 |
| RMSE | 9.449±1.065 | 8.101±0.117 | 8.808±0.737 | 9.645±0.884 |
| WMAPE | 0.636±0.074 | 0.439±0.006 | 0.494±0.043 | 0.660±0.083 |

Table V shows the performance of our model with different dimension numbers of LSTM input, i.e., $hid_2$. The result shows the best size of $hid_2$ is 24. The reason might be that the smaller dimension number of input loses information, but bigger dimension number of input introduces more parameters and is prone to overfitting due to limited training data.

### 4.6. Relative error analysis

To analyze the prediction versus the ground truth, we visualize the data for a total of three days from January 17, 2021 to January 19, 2021 in the test set. Each point represents the infection number of a community at a day.

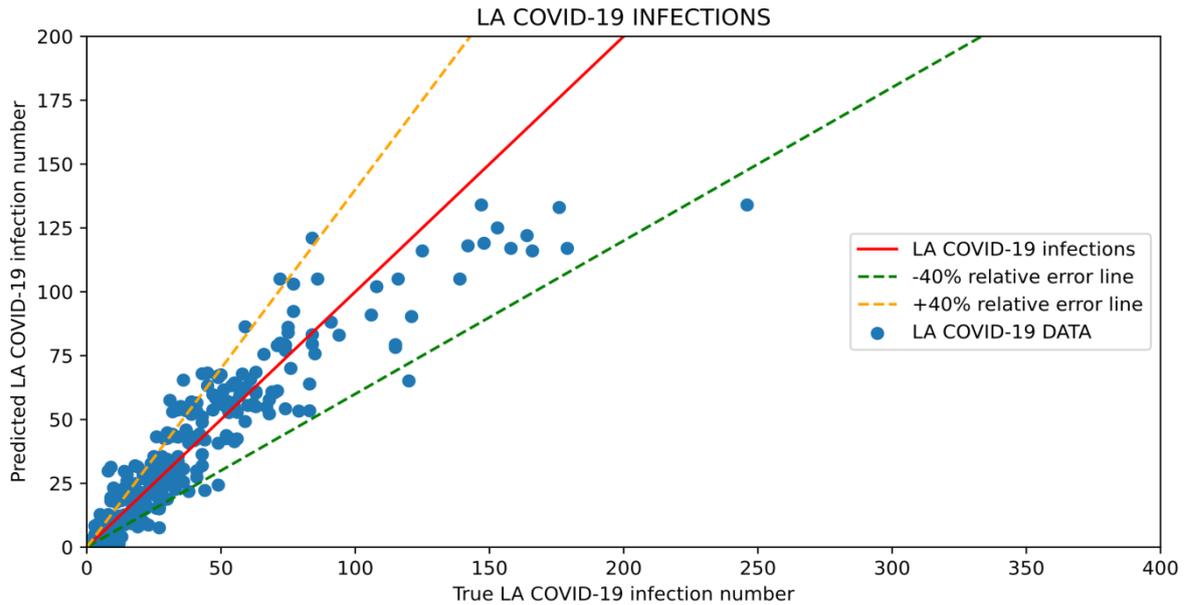

Figure 5: The relative error between the true and the predicted infection numbers

Figure 5 shows the relative error between the true and the predicted infection numbers. It can be seen that, most of the data points fall within the 40% relative error band. Due to the specificity of the epidemic data, numerous data points are clustered around the value of 0. Considering the existence of recording errors in the epidemic data and the presence of many small values, the prediction performance of FGC-COVID is remarkable.

### 4.7. Case analysis

We visualize the ground truth and the predictions of infections made by FGC-COVID and other better-performing models at the Downtown community.

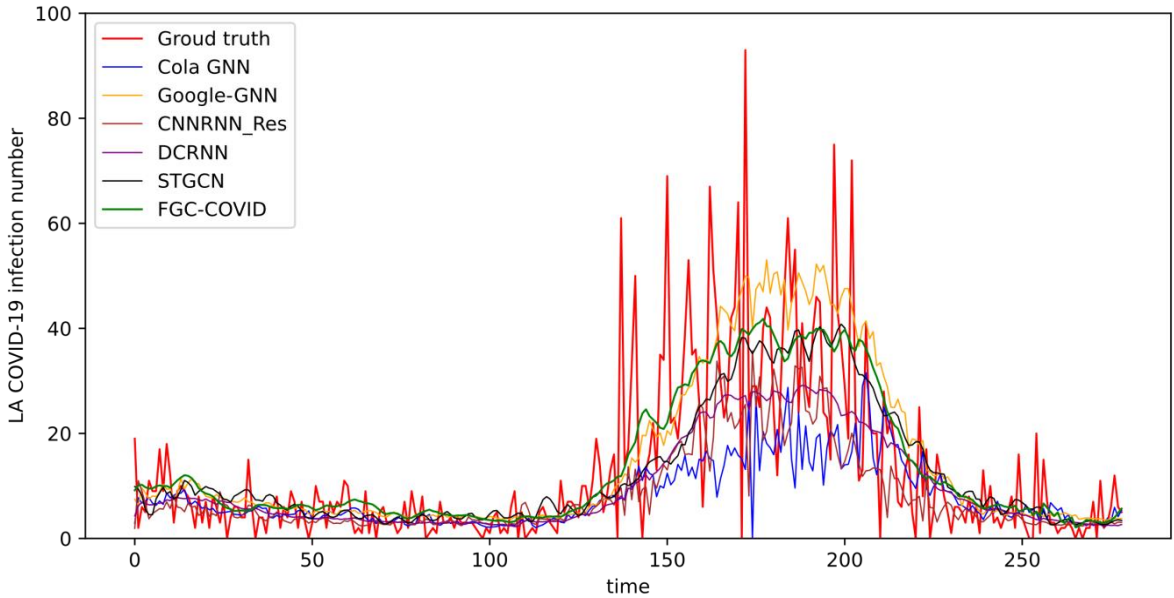

Figure 6: The visualization of the true and the predicted infection numbers at the Downtown community

From Figure 6, we can see that FGC-COVID fits the data best. For the zero values that appear during the peak of the epidemic, they may be due to the oversight in the data statistics. In terms of overall trend, FGC-COVID can capture the data fluctuation well, and it fits and predicts the COVID-19 data better than other baselines. In addition, our model can better fit the fluctuation of the outbreak stage that appears in the range of the 150th day to the 200th day. The fitting curve of FGC-COVID is basically consistent with the rising and falling trend of the real data, and at the same time, the fitting value is also controlled within a reasonable interval. The reason might be that we extract the spatial pattern at a lower geographic level that has more geographic units, which helps the model capture the fluctuations in epidemic data.

## 5. Conclusions and Future Work

In this paper, we focus on community level COVID-19 infection prediction problem and propose FGC-COVID. We introduce a spatial pattern extraction module to capture the spatial pattern on CBG-level, a spatial weighted aggregation module to aggregate the embeddings of CBGs based on their geographic affiliation to community level, and use LSTM to extract temporal patterns. We also compare FGC-COVID with SOTA baselines on the COVID-19 dataset from LA city. The results show that our model has greater performance in terms of prediction error.

In the future, we will deepen our work in the following aspects: (1) introduce a multi-level geographic structure in the model to capture more complex correlations between CBGs and communities and (2) introduce more related data, e.g., vaccination data.

# 6. References

Alzahrani, S. I., I. A. Aljamaan, and E. A. Al-Fakih. 2020. Forecasting the spread of the COVID-19 pandemic in Saudi Arabia using ARIMA prediction model under current public health interventions. *Journal of Infection and Public Health* 13 (7):914-19.

Anselin, L. 1995. Local indicators of spatial association-LISA. *Geographical Analysis* 27 (2):93-115.

Arora, P., H. Kumar, and B. K. Panigrahi. 2020. Prediction and analysis of COVID-19 positive cases using deep learning models: A descriptive case study of India. *Chaos, Solitons & Fractals* 139:110017.

Banerjee, S., and Y. Lian. 2022. Data driven COVID-19 spread prediction based on mobility and mask mandate information. *Applied Intelligence* 52 (2):1969-78.

Calafiore, G. C., C. Novara, and C. Possieri. 2020. A modified SIR model for the COVID-19 contagion in Italy. In *Proceedings of the 59th IEEE Conference on Decision and Control*, 3889-94.

Ceylan, Z. 2020. Estimation of COVID-19 prevalence in Italy, Spain, and France. *Science of The Total Environment* 729:138817.

Cuevas, E. 2020. An agent-based model to evaluate the COVID-19 transmission risks in facilities. *Computers in Biology and Medicine* 121:103827.

Deng, S., S. Wang, H. Rangwala, L. Wang, and Y. Ning. 2020. Cola-GNN: Cross-location attention based graph neural networks for long-term ILI prediction. In *Proceedings of the 29th ACM International Conference on Information & Knowledge Management*, 245-54.

Gallagher, S., and J. Baltimore. 2017. Comparing compartment and agent-based models. In *Proceedings of the Joint Statistical Meeting*. Alexandria, VA: American Statistical Association.

He, S., Y. Peng, and K. Sun. 2020. SEIR modeling of the COVID-19 and its dynamics. *Nonlinear Dynamics* 101 (3):1667-80.

Jing, N., Z. Shi, Y. Hu, and J. Yuan. 2022. Cross-sectional analysis and data-driven forecasting of confirmed COVID-19 cases. *Applied Intelligence* 52 (3):3303-18.

Kapoor, A., X. Ben, L. Liu, B. Perozzi, M. Barnes, M. Blais, and S. O'Banion. 2020. Examining COVID-19 forecasting using spatio-temporal graph neural networks. *arXiv preprint arXiv:2007.03113*.

Kiamari, M., G. Ramachandran, Q. Nguyen, E. Pereira, J. Holm, and B. Krishnamachari. 2020. COVID-19 risk estimation using a time-varying SIR-model. In *Proceedings of the 1st ACM SIGSPATIAL International Workshop on Modeling and Understanding the Spread of COVID-19*, 36-42.

Kipf, T. N., and M. Welling. 2017. Semi-supervised classification with graph convolutional networks. Poster presented at International Conference on Learning Representations, Palais des Congrès Neptune, Toulon, France, April 26.

Lai, G., W.-C. Chang, Y. Yang, and H. Liu. 2018. Modeling long-and short-term temporal patterns with deep neural networks. In *Proceedings of the 41st International ACM SIGIR Conference on Research & Development in Information Retrieval*, 95-104.

Li, Y., R. Yu, C. Shahabi, and Y. Liu. 2018. Diffusion convolutional recurrent neural network: Data-driven traffic forecasting. Poster presented at International Conference on Learning Representations, Vancouver, BC, Canada, April 30.

Loshchilov I., and F. Hutter. 2019. Decoupled weight decay regularization. Poster presented at International Conference on Learning Representations, New Orleans, Louisiana, May 8.

Panagopoulos, G., G. Nikolentzos, and M. Vazirgiannis. 2021. Transfer graph neural networks for pandemic forecasting. In *Proceedings of the 35th AAAI Conference on Artificial Intelligence*, 4838-45.

Pandey, G., P. Chaudhary, R. Gupta, and S. Pal. 2020. SEIR and regression model based COVID-19 outbreak predictions in India. *arXiv preprint arXiv:2004. 00958*.

Rodrǵuez, A., N. Muralidhar, B. Adhikari, A. Tabassum, N. Ramakrishnan, and B. A. Prakash. 2021. Steering a historical disease forecasting model under a pandemic: Case of flu and COVID-19. In *Proceedings of the 35th AAAI Conference on Artificial Intelligence*, 4855-63.

Roy, S., G. S. Bhunia, and P. K. Shit. 2021. Spatial prediction of COVID-19 epidemic using ARIMA techniques in India. *Modeling Earth Systems and Environment* 7 (2):1385-91.

Shamil, M. S., F. Farheen, N. Ibtehaz, I. M. Khan, and M. S. Rahman. 2021. An agent-based modeling of COVID-19: Validation, analysis, and recommendations. *Cognitive Computation* 1-12.

Singh, R. K., M. Rani, A. S. Bhagavathula, R. Sah, A. J. RodriguezMorales, H. Kalita, C. Nanda, S. Sharma, Y. D. Sharma, A. A. Rabaan, et al. 2020. Prediction of the COVID-19 pandemic for the top 15 affected countries: Advanced autoregressive integrated moving average (ARIMA) model. *JMIR Public Health and Surveillance* 6 (2):19115.

Wang, P., X. Zheng, G. Ai, D. Liu, and B. Zhu. 2020. Time series prediction for the epidemic trends of COVID-19 using the improved LSTM deep learning method: Case studies in Russia, Peru and Iran. *Chaos, Solitons & Fractals* 140:110214.

Wu, Y., Y. Yang, H. Nishiura, and M. Saitoh. 2018. Deep learning for epidemiological predictions. In *Proceedings of the 41st International ACM SIGIR Conference on Research & Development in Information Retrieval*, 1085-88.

Yu, B., H. Yin, and Z. Zhu. 2018. Spatio-temporal graph convolutional networks: A deep learning framework for traffic forecasting. In *Proceedings of the 27th International Joint Conference on Artificial Intelligence*, 3634-40.

Zeroual, A., F. Harrou, A. Dairi, and Y. Sun. 2020. Deep learning methods for forecasting COVID-19 time-series data: A comparative study. *Chaos, Solitons & Fractals* 140:110121.